\documentclass{article}
\usepackage[utf8]{inputenc}
\usepackage[T1]{fontenc}
\usepackage{url}
\usepackage{tikz}
\usepackage{tabularx}
\usepackage{multirow}
\usepackage{booktabs}
\usepackage{listings}

\usepackage{doc}

\title{Detecting New Word Meanings: A Comparison of Word Embedding Models in Spanish}

\author{
Andrés Torres-Rivera $^{1,3}$ and Juan-Manuel Torres-Moreno$^{2,3}$\\
   $^1$ Univrersitat Pompeu Fabra,\\
  Roc Boronat 138 Barcelona, Spain\\
  {andres.torres@upf.edu}\\
   $^2$ Polytechnique Montr\'eal,\\ CP. 6128 succursale Centre-ville,
   Montr\'eal (Québec) Canada\\
  $^3$ Laboratoire Informatique d'Avignon,\\
  BP 91228 84911, 
  Avignon, Cedex 09, France\\
  {juan-manuel.torres@univ-avignon.fr}
 }

\begin{document}

\maketitle

\begin{abstract}
Semantic neologisms (SN) are defined as words that acquire a new word meaning while maintaining their form. Given the nature of this kind of neologisms, the task of identifying these new word meanings is currently performed manually by specialists at observatories of neology. To detect SN in a semi-automatic way, we developed a system that implements a combination of the following strategies: topic modeling, keyword extraction, and word sense disambiguation.

The role of topic modeling is to detect the themes that are treated in the input text. Themes within a text give clues about the particular meaning of the words that are used, for example: viral has one meaning in the context of computer science (CS) and another when talking about health. To extract keywords, we used TextRank with POS tag filtering. With this method, we can obtain relevant words that are already part of the Spanish lexicon. We use a deep learning model to determine if a given keyword could have a new meaning. Embeddings that are different from all the known meanings (or topics) indicate that a word might be a valid SN candidate.
In this study, we examine the following word embedding models: Word2Vec, Sense2Vec, and FastText. The models were trained with equivalent parameters using Wikipedia in Spanish as corpora. Then we used a list of words and their concordances (obtained from our database of neologisms) to show the different embeddings that each model yields. Finally, we present a comparison of these outcomes with the concordances of each word to show how we can determine if a word could be a valid candidate for SN.
\end{abstract}

%------------------------------------------------------------------------------
\section{Introduction}

%The detection of semantic neologisms (SN) is a task that is currently performed by specialists at observatories of neology such as OBNEO\footnote{Observatori de Neologia (OBNEO) \url{https://www.upf.edu/web/obneo/}}, NEOPORTERM\footnote{Observatório de Neologia e de Terminologia em Língua Portuguesa (NEOPORTERM) \url{http://clunl.fcsh.unl.pt/investigacao/projetos-concluidos/neoporterm-observatorio-de-neologia-e-de-terminologia-em-lingua-portuguesa/}} and OBNEQ\footnote{Observatoire de Néologie du Québec (OBNEQ) \url{http://www.lli.ulaval.ca/recherche/groupes-et-laboratoires}}. This task usually consists in reading and analyzing newspapers to highlight ambiguous words that might be valid candidates for SN. A valid candidate is defined as a known word that, in a certain context, acquires a new meaning, usually a meaning related to the subject that is being treated. For further delimitation of this phenomenon we refer to the classification of neologisms proposed by \cite{cabre2009}: a semantic neologisms is a type of neologism that goes trough a semantic process which could reduce, expand, or change its meaning.

\begin{table}[h]{Excerpt of the multivariable table \cite[p. 35]{cabre2009}}
\small
\centering
\begin{tabular}{lllll}
\toprule
\multirow{5}{*}{\textbf{Process}} & \multirow{5}{*}{\textbf{Formation}} & \multirow{5}{*}{\textbf{\begin{tabular}[c]{@{}l@{}}Grammatical\\ change\end{tabular}}} & \multicolumn{2}{l}{\textbf{Category (FCONV)}}\\ \cline{4-5} 
&&& \multicolumn{2}{l}{\textbf{Subcategory (SINT)}}\\ \cline{4-5} 
&&& \multirow{3}{*}{\textbf{\begin{tabular}[c]{@{}l@{}}Semantic\\ (re-\\semantization)\end{tabular}}}	& Meaning reduction  \\  
&&&& Meaning expansion \\
&&&& Meaning change    \\
\bottomrule
\end{tabular}
\label{tabla:cabre2009}
\end{table}

The scope of this study is limited to SN that have their origin as specialized knowledge units (terms) as defined by \cite{Cabre2005}:

\begin{quote}
    [...]a lexical unit, whose structure is related to an origin lexical unit or product of the lexicalization of a syntagm, that has a specific meaning in the ambit to which it is related and it is necessary in the conceptual structure of the domain in which it takes part. \cite[p.77]{Cabre2005}
\end{quote}

These terminological units are part of a broader general concept defined as specialized significance units (SSU\footnote{In Catalan: \textit{Unitats de Significació Especializada.}}). There are different types of SSU, such as lexical, verbal, nominal, adjectival or adverbial \cite{Estopa2013}. In this study we will explore verbal, nominal and adjectival SSU obtained from our SN database. To carry out the detection of SN in a semiautomatic way, we designed a system that uses topic detection, keyword extraction and word embeddings to detect new word meanings. We named this system DENISE\footnote{A play on words that fits its three working languages: \textit{detector de neologismos semánticos}, in Spanish; \textit{detector de neologismes semàntics}, in Catalan and \textit{détecteur de néologismes sémantiques} in French.}, a multilingual tool that mimics the work that specialists perform at OBNEO. 

DENISE takes simple text as input, then the user can set the working language or let the system detect it automatically, currently the supported languages are: Catalan, French and Spanish. The next step is text preprocessing: normalization and tokenization as well as elimination of stopwords, non latin characters and symbols. Once the text is preprocessed the user has the option to introduce the theme of the text or let the system detect the topics within the text using a TF-IDF and logistic regression based model. Once the text has a topic assigned, the system will automatically extract keywords using TextRank with POS filtering.

\begin{figure}[h]
    \centering
    \includegraphics[width=1\columnwidth]{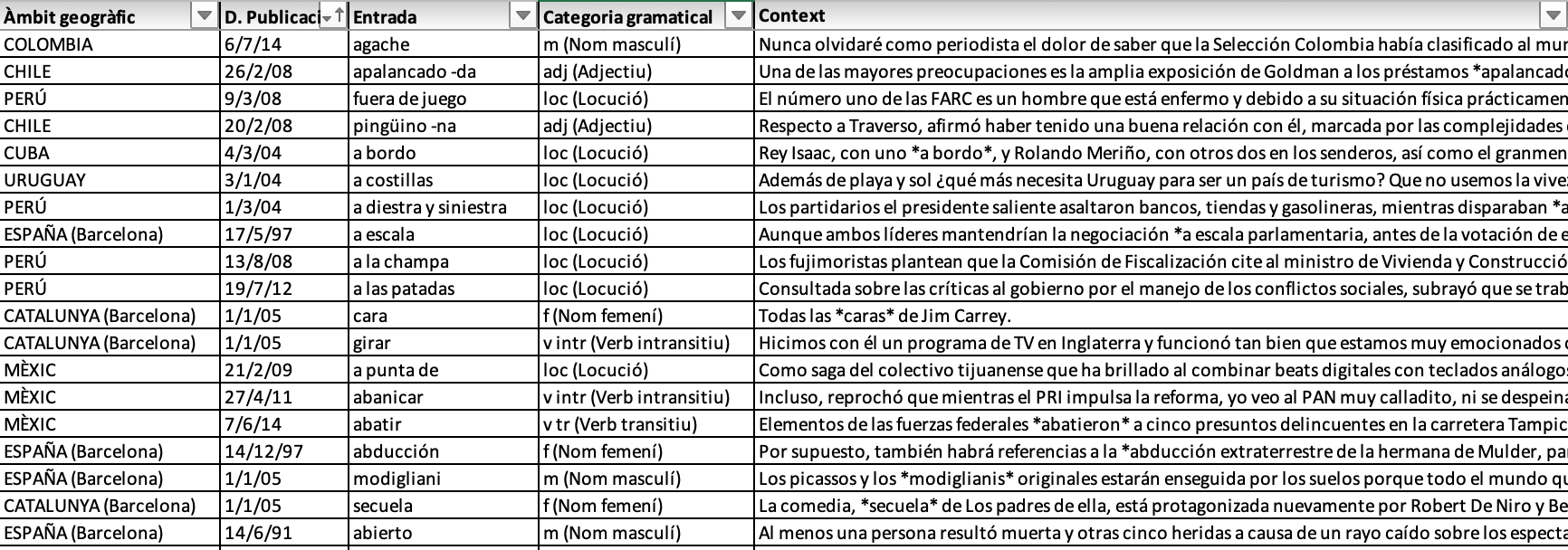}
    \caption{Excerpt from the OBNEO Database}
    \label{fig:obneo_db}
\end{figure}

Each of these keywords serve as query items for our word embeddings models; with these queries we can obtain the 140 most similar terms\footnote{We set topn=140 because the input texts are short concordances, varying from 130 to 150 characters each after filtering concordances with non-informative contexts.} and then create a "semantic field" (SF) for each keyword. This SF serves as a representation of the most common meaning of each queried word in the general language. Then, the system evaluates these SF to detect their theme and proceeds to evaluate if concordance exists between the topic of the input text and the topic of the SF. To obtain valid candidates for SN, DENISE filters the keywords keeping only keywords that meet the following criteria: a candidate for SN is a keyword whose detected topic is different from the topic of the embeddings, since this would indicate that a known word is being used is a context that is different from its most common SF.

%------------------------------------------------------------------------------

%------------------------------------------------------------------------------

\section{Related Work}

The automatic detection of SN, because of its nature, has been a more complicated task compared with other kinds of neologisms such as transferred words or derived words \cite{Tebe2002, Janssen2009, Renouf2010, Sablayrolles2012, Reutenauer2011}. Therefore, presently there is a necessity that is being covered partially. One of the first specialized systems to detect SN is April \cite{Renouf1998, Renouf2010, Renouf2012}; this approach uses statistical and linguistic rules, collocation patterns and heuristic rules to track semantic change over time. These methods include boot-strapping, a chronologically divided corpus used as a control database and a reference dictionary. April analyzes common collocations to track SN, which means that a word that a appears in a context different from its most usual context might be a valid candidate for SN. However, no evaluation is provided and the authors mention two specific problems: the superficial definition of novelty does not distinguish between a new sense or a new reference, and April can not identify collocations that have at least four concordances.

A second system that could be used to detect SN is Logoscope\footnote{\url{http://logoscope.unistra.fr}} \cite{Falk2014-1, Falk2014-2, Falk2014-3, Gerard2014, Bernhard2015, Bernhard2015a}, which uses a combination of topic modeling using Latent Dirichlet Allocation (LDA) and a linear support vector machine classifier that could be used to identify possible new word meanings. In order to detect SN, Logoscope analyzes theme concordance between the collocation of a keyword and its definition in the dictionary: when the collocation and the definition do not share the same topic this could indicate that a given keyword might be candidate for SN. The authors were able to detect a new sense for the word \textit{quenelle} using this methodology, but they affirm that relying on dictionary definitions complicates this task given the nature of SN.  While the authors provide all the formulas that were used, there is no formal evaluation for this particular task.

Finally, there are other methodological approaches such as those proposed by \cite{Janssen2005, Janssen2009, Janssen2012}, who developed a POS tagger that uses statistical rules and could obtain SN as a byproduct from this process: ambiguous words are assigned a special tag that, when inspected, can indicate a semantic change in course. Finally, \cite{Nazar2011, Nazar2013, Nazar2014} proposes a methodology that involves the combination of word sense induction (WSI) and clustering to group word senses. The author states that this method could be implemented to develop a tool that detects SN automatically. While both methodologies explain how word meanings can be grouped and classified, neither provide implementations for the detection of SN.

\section{Theme Detection}

As part of our methodology, we expect to classify the different themes or topics that are being treated in the input text, because we assume that new word meaning depends on the topic in which we find this word. For example, if a word has one known meaning related to economy and we find this same word in a CS text, this word might have a change of meaning related to this new theme. We treated this step as a classification problem, which means that each text that is entered to DENISE is evaluated using a logistic regression to predict its main theme or topic. 

\begin{figure}[h]
    \centering
    \includegraphics[width=0.6\columnwidth]{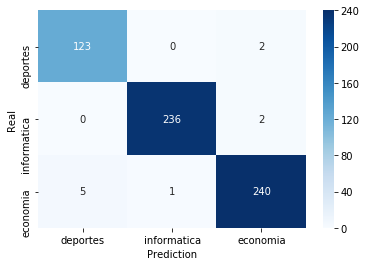}
    \label{fig:model-comp}
    \caption{Confusion Matrix of the Logistic Regression Model}
\end{figure}

Our corpus was compiled using articles from specialized publications in Spanish: PC World (CS) with a total of 308,930 words; Marca (sports), 275,872 words; and El Financiero (economy), 280,404 words. This corpus was used to generate a TF-IDF model and then train the logistic regression model using the following parameters: L2 penalty, max intercept scaling of 1, max tolerance of 1e-4 and 1.0 for inverse of regularization strength, the resulting confusion matrix can be seen in Figure \ref{fig:model-comp}.

The model is capable of detecting three topics: sports (\textit{de\-por\-tes}), economy (\textit{e\-co\-no\-mí\-a}), and computer science (\textit{in\-for\-má\-ti\-ca}). As we mentioned before, our goal is to detect new word meanings related to computer science, therefore we expect that, for every text we evaluate, DENISE will return ``informática'' (labeled as 1) as the main topic, and if it does not detect CS as the main topic it will return ``not informática'' (labeled as 0). We compared the precision of three classifiers using a cross validation score: logistic regression obtained 0.982; multinomial naïve Bayes, 0.898; and random forest classifier, 0.790. After obtaining these results, we selected the logistic regression model and proceeded to evaluate its mean accuracy on a train-test split: it obtained 0.913 on the train set and 0.889 on the test set.

\section{Keyword Extraction}

After detecting the theme of the input text, DENISE extracts the keywords that will be evaluated as candidates. One of the particularities of SN is that they are known words and, therefore, we can not use a set of lexicographical rules or dictionaries because we are not looking for new words at a formal (structural) level, but for new meaning of known words. For this reason we decided to use the TextRank \cite{Mihalcea2004} algorithm (as defined by Equation \ref{eq:textrank}) with POS tag filtering. This graph-based algorithm was inspired by the PageRank \cite{Page1999} algorithm originally used by the Google search engine. TextRank is a widely used in ranking and recommendation systems, keyword extraction and automatic summarization systems \cite{Li2014, Barrios2016, Pay2018}.

\begin{equation}
    \label{eq:textrank}
    WS(V_i) = (1-d) + d \times \sum_{V_j \in In (V_i) } \frac{w_{ij}}{\sum_{V_k \in Out (V_j) } w_{jk}} WS(V_j)
\end{equation}

DENISE's implementation of TextRank uses the original graph evaluation and incorporates POS tag filtering to prioritize the extraction of verbs, nouns and adjectives. These type of units are of interest because in our database of neologisms we found that most SN, as shown in Table \ref{tab:sn_by_pos}, fall into these POS categories. With the implementation of a POS filter we obtained 14\% more accuracy in comparison with the regular TextRank implementation, this might indicate that the algorithm is correctly extracting possible candidates.

\begin{table}[h]{NS Distribution by Part of Speech}
\small
\label{tab:sn_by_pos}
\centering
%\small\addtolength{\tabcolsep}{-5pt}
\begin{tabular}{lr}
\toprule
POS         &   Total\\
\midrule
adj (Adjective)               &    750 \\
adv (Adverb)                &      3 \\
conj (Conjunction)             &      2 \\
f (Feminine Noun)               &   1280 \\
f pl (Feminine Plural Noun)     &     20 \\
interj (Interjection)         &      3 \\
loc (Locution)                &     85 \\
m (Masculine Noun)              &   2425 \\
m i f (F. and M. Noun) &     61 \\
m pl (Plural Masculine Noun)    &     62 \\
n (Neutral Noun)               &      3 \\
pron (Pronoun)                &      2 \\
v intr (Intransitive Verb)    &    146 \\
v pron (Pronominal Verb)     &    129 \\
v tr (Transitive Verb)        &    591 \\
\bottomrule
\end{tabular}
\end{table}

\section{Word Embedding Models}

The last step in DENISE's analysis process is sense disambiguation using word embeddings. With the resulting keyword from the previous step, our work hypothesis is the following: a new word sense might be found when a known word is used in a text about a topic that is different from the topics where this word is usually collocated. Therefore we assume that the most common word representation of a given keyword is closely related to its main (or most common) meaning, and this meaning might is also related to a certain topic.

We carried out an analysis using three different neural network based models: Word\-2\-Vec \cite{Mikolov2013, Mikolov2013b}, FastText \cite{Bojanowski2016, Joulin2016} and Sense2Vec \cite{Trask2015}. To train the models we used Wikipedia in Spanish as corpus and the training was performed using the same training values described in the bibliography. In the following subsections we describe some of the particularities of each model and the training parameters that were used.

\begin{description}
\item{\textbf{Word2Vec}}. 
Word2Vec is a model that uses neural networks to produce dense vector representations of words. Its two main architectures are skip-gram and CBOW, the first one being slower but better for projecting uncommon words. In order to perform the disambiguation process, we require embeddings that represent the most common meaning of the input keywords, therefore we trained our skip-gram based model with a dimension size of 300, a window of 5 and a min count of 20 elements.

\item{\textbf{Sense2Vec}}. 
To obtain a Sense2Vec model we tagged the same Wikipedia corpus that was used to train the Word2Vec model, using the Universal Dependencies\footnote{\url{http://universaldependencies.org/u/pos/}} tagset. 
This approach allows for the generation of a model that has similar characteristics to a Word2Vec model with the added advantage of POS tag disambiguation. This means that a word that is being used as a noun and a verb has two different representations in the model, one for each case. We followed the training parameters proposed by \cite{Trask2015}: a dimension size of 500, a window of 5 and a min count of 10 elements, again using continuous skip-grams.

\item{\textbf{FastText}}.
Altough the FastText model also uses neural networks to generate word representations, this model uses subword data as the minimum units to train these representations;
 meanwhile, the minimum unit to trained by Word2Vec and Sense2Vec are words. For example, in a FastText model the vector for the word "viral" would be composed by the ngrams within "viral" in the following way: "<vi", "vir", "vira", "viral", "viral>", "ira", "iral>", "iral>", "ral", "ral>", "al>". We used the pretrained vectors available at the FastText website\footnote{\url{https://fasttext.cc/docs/en/crawl-vectors.html}}, which were trained using CBOW with position-weights, in dimension 300, with character n-grams of length 5, a window of size 5 and 10 negatives.
\end{description}

\section{Pipeline Description}

To extract SN from a text first our algorithm (\ref{tab:pseudocodigo}) detects the working language, if the language is not supported it continues with a new text. If the language is supported then DENISE proceeds to assign a theme to the input text using the logistic regression model. If the user knows the language and the topic it can be set manually, the number of keywords to extract and the number of words for the SF can also be set manually.

The next step is to extract keywords using TextRank, the implementation returns a list with the $n$ number of KW that were selected by the user. For each of the items from the TextRAnk list, the algorithm returns a SF composed by the TOPN most similar words for each item of the list using the word embedding model, which by default is set to Sense2Vec.

Finally, the systems evaluates if there is theme concordance between the semantic field and the input text. When concordance exists it indicates that the most common meaning from the embedding model is being use, which means that a given word is not a candidate for SN. Meanwhile, if concordance does not exist between both topic, this might indicate that a given keyword is a possible candidate for SN, when this is the case, the algorithm returns a list with the possible candidates and their detect themes so the user can determine if there are true SN candidates.

\begin{table}[!h]{Pseudocode Description}
\lstset{basicstyle=\footnotesize}
\begin{lstlisting}
def sn_classification(text,lang='auto',topic='auto',KW=10,TOPN=140):
    if lang == 'auto':
        lang_flag = lang_detection(text)
    else:
        lang_flag = True, lang
    
    if lang_flag[0] == True:
        sn_list = []
        if topic == 'auto':
            text_topic = analyze_topic([text], lang_flag[1])
        else:
            text_topic = topic
        text_rank = extract_keywords(text, lang_flag[1], KW)
        for i in text_rank:
            sf = model.most_similar(i, TOPN)
            sf_topic = analyze_topic([sf], lang_flag[1])
            if text_topic != sf_topic:
                sn_list.append([i, text_topic, sf_topic])
            else:
                continue
        return sn_list
    else:
        print("Language not supported")
\end{lstlisting}
    \label{tab:pseudocodigo}
\end{table}

\section{Model Comparison}

To determine the performance of DENISE with each model, we used a list of keywords and their concordances (all pertaining to the CS field) from the SN database as input text. Our database has a total of 5562 SN registered and manually validated; from this total we selected SN that correspond to the CS field and discarded concordances shorter than 130 words to have enough context words. These are SN that have already been manually analyzed and all of them belong to the field of the CS.

With these keywords and concordances we generated a CSV table containing the 125 term-concordance items, then we proceeded to query each item of the list on each model to obtain the 140 most similar terms for each item. As shown in Table \ref{tab:model_query}, FastText was the model that yielded the best results, since with this model we obtained 125 SF for each of the 125 terms that were queried, followed by the Word2Vec model with 100 from 125 terms, and last the Sense2Vec model with 97 of 125.

\begin{table}[!htb]{Items obtained for each model}
\small
    \centering
    \begin{tabular}{lrrr}
    \toprule
Model & Expected & Recovered & Percentage\\
\midrule
FastText & 125 & 125 & 100\% \\
Word2Vec & 125 & 100 & 80\% \\
Sense2Vec & 125 & 97 & 77.6\% \\
\bottomrule
    \end{tabular}
    \label{tab:model_query}
\end{table}

After obtaining the most similar terms for each item, we performed the topic detection process on these SF. This process is performed in order to assess if the generated SF have the same topic than the concordances. If the set of embeddings and the concordance share the same topic, it might indicate that the keyword in question might no be candidate for SN, whereas a set of dissimilar embeddings and concordance might indicate that the keyword is a candidate for SN. The results of this process are shown in Tables \ref{tab:model_eval_w2v_ft} and \ref{tab:model_eval_s2v}.

\begin{table}[!htb]{Results of FastText and Word2Vec Models}
\small
    \centering
    \begin{tabular}{lrrrr}
    \toprule
Model & SF & CS Label & Man. Eval. & Total Correct \\
\midrule
FastText & 125 & 38 & 78 & 77 \\
Word2Vec & 100 & 34 & 57 & 63 \\
\bottomrule
    \end{tabular}
    \label{tab:model_eval_w2v_ft}
\end{table}

While the Sense2Vec model was the model that retrieved the lowest number of unique terms, it provided the more representations in total: 172 when combining all three categories. This gives us more detailed information regarding use: a word might be used mainly as a verb but it could also be used a noun, and one of these uses could be the neological meaning.

\begin{table}[!htb]{Results of the Sense2Vec Model}
\small
    \centering
    \begin{tabular}{lrrrr}
    \toprule
Tag & SF & CS Label & Man. Eval. & Total Correct \\
\midrule
VERB & 35 & 7 & 15 & 25 \\
NOUN & 87 & 32 & 44 & 73 \\
ADJ & 50 & 17 & 26 & 41 \\
\bottomrule
    \end{tabular}
    \label{tab:model_eval_s2v}
\end{table}

We evaluated each of the SF manually to ensure that the automatic topic detection process was accurate, and to ensure that the model classifies the terms correctly: whether they belong to CS or not. After manually evaluating all the embeddings, we proceeded to evaluate the number of correct cases, that is, if the predicted topic is the same as the manually observed topic and the percentage of agreement between the automatically and the manually labeled SF. We expected that the classifier could determine if a context and a SF are, in fact, related to the CS field or not. We used f1-Score, precision, recall and support for each model; in the case of the Sense2Vec model we calculated these metric for each POS category independently.

The results for the FastText model can be seen in Table \ref{tab:ft_eval} and the results for Word2Vec on Table \ref{tab:w2v_eval}. Both models obtained similar f1-scores, and both seem to have high precision and low recall when classifying SF that do not belong to CS, but also present low precision and high recall when classifying SF that belong to CS.

\begin{table}[h]{FastText classification results}
\small
    \centering
    \begin{tabular}{lrrrr}
    \toprule
                 &  f1-score &  precision &    recall &  support \\
    \midrule
    0            &  0.64179 &   0.91489 &  0.49425 &     87.0 \\
    1            &  0.58621 &   0.43590 &  0.89474 &     38.0 \\
    micro avg    &  0.61600 &   0.61600 &  0.61600 &    125.0 \\
    macro avg    &  0.61400 &   0.67540 &  0.69450 &    125.0 \\
    weighted avg &  0.62489 &   0.76928 &  0.61600 &    125.0 \\
    \bottomrule
    \end{tabular}
    \label{tab:ft_eval}
\end{table}

\begin{table}[h]{Word2Vec classification results}
\small
    \centering
    \begin{tabular}{lrrrr}
    \toprule
                 &  f1-score &  precision &    recall &  support \\
    \midrule
    0            &  0.66055 &   0.83721 &  0.54546 &     66.0 \\
    1            &  0.59341 &   0.47368 &  0.79412 &     34.0 \\
    micro avg    &  0.63000 &   0.63000 &  0.63000 &    100.0 \\
    macro avg    &  0.62698 &   0.65545 &  0.66979 &    100.0 \\
    weighted avg &  0.63772 &   0.71361 &  0.63000 &    100.0 \\
    \bottomrule
    \end{tabular}
    \label{tab:w2v_eval}
\end{table}

On the other hand, the Sense2Vec model obtained better f1-scores than the Word2Vec and FastText model, with being NOUN the most productive --and balanced-- category with a weighted average f1-score of 0.84. This value is also greater than both the f1-score of the Word2Vec model and the FastText model. Overall, while the Sense2Vec model retrieved the least amount of SF, the resulting embeddings were better classified.

\begin{table}[h]{Sense2Vec classification results}
\small
    \centering
    \begin{tabular}{lrrrr}
    \toprule
    \multicolumn{5}{c}{Verbs}\\
                 &  f1-score &  precision &    recall &  support \\
    \midrule
    0            &  0.79167 &   0.95000 &  0.67857 &     28.0 \\
    1            &  0.54546 &   0.40000 &  0.85714 &      7.0 \\
    micro avg    &  0.71429 &   0.71429 &  0.71429 &     35.0 \\
    macro avg    &  0.66856 &   0.67500 &  0.76786 &     35.0 \\
    weighted avg &  0.74242 &   0.84000 &  0.71429 &     35.0 \\
    \midrule
    \multicolumn{5}{c}{Nouns}\\
                 &  f1-score &  precision &    recall &  support \\
    \midrule
    0            &  0.85714 &   0.97674 &  0.76364 &     55.0 \\
    1            &  0.81579 &   0.70455 &  0.96875 &     32.0 \\
    micro avg    &  0.83908 &   0.83908 &  0.83908 &     87.0 \\
    macro avg    &  0.83647 &   0.84065 &  0.86619 &     87.0 \\
    weighted avg &  0.84193 &   0.87663 &  0.83908 &     87.0 \\
    \midrule
    \multicolumn{5}{c}{Adjectives}\\
                 &  f1-score &  precision &    recall &  support \\
    \midrule
    0            &  0.84211 &   1.00000 &  0.72727 &     33.0 \\
    1            &  0.79070 &   0.65385 &  1.00000 &     17.0 \\
    micro avg    &  0.82000 &   0.82000 &  0.82000 &     50.0 \\
    macro avg    &  0.81640 &   0.82692 &  0.86364 &     50.0 \\
    weighted avg &  0.82463 &   0.88231 &  0.82000 &     50.0 \\
    \bottomrule
    \end{tabular}
    \label{tab:s2v_eval}
\end{table}

Finally, following the condition of disagreement between the topic of the embeddings and the topic of the input text, each model generated a list o candidates for SN. The Sense2Vec model generated a list of 55 candidates from the original 125 SN list, FastText 42 candidates and Word2Vec, 35 candidates. The lists that each model generated are shown below:

\begin{quote}
\small
    \textbf{Sense2Vec Candidates:} 'almacenado', 'navegabilidad', 'palm', 'mini', 'cablear', 'controladora', 'terminal', 'viral', 'descarga', 'navegación', 'cargarse', 'objeto', 'cuenta', 'perfil', 'visual', 'directorio', 'asistente', 'bitácora', 'acelerar', 'chip', 'caída', 'caerse', 'conversión', 'muro', 'word', 'cortafuego', 'vacuna', 'nube', 'infectar', 'celular', 'gusano', 'troyano', 'dominio', 'navegar', 'alojamiento', 'electrónico', 'portal', 'migración', 'aplicación', 'ipod', 'motor', 'procesador', 'agujero', 'avatar', 'androide', 'piratería', 'virus', 'enlace', 'apuntador', 'subir', 'clo\-na\-ción', 'vínculo', 'api', 'herramienta', 'guru'.

    \textbf{FastText Candidates:} 'almacenado', 'navegabilidad', 'palm', 'jaquear', 'controladora', 'game boy', 'navegación', 'cargarse', 'iserie', 'cuenta', 'clonar', 'visual', 'menú', 'asistente', 'acelerar', 'caída', 'caerse', 'conversión', 'mapeo', 'muro', 'vacuna', 'nube', 'infectar', 'gusano', 'dominio', 'navegar', 'alojamiento', 'correo\-\_electrónico', 'electrónico', 'migración', 'clic', 'motor', 'agujero', 'avatar', 'virus', 'apuntador', 'subir', 'vínculo', 'disco duro', 'descargar', 'guru', 'descargarse'.

    \textbf{Word2Vec Candidates:} 'almacenado', 'navegabilidad', 'palm', 'jaquear', 'viral', 'parche', 'navegación', 'cargarse', 'cuenta', 'clonar', 'visual', 'asistente', 'acelerar', 'caída', 'caerse', 'conversión', 'muro', 'word', 'vacuna', 'nube', 'infectar', 'troyano', 'gusano', 'dominio', 'navegar', 'alojamiento', 'migración', 'motor', 'agujero', 'avatar', 'piratería', 'virus', 'subir', 'vínculo', 'guru'.
\end{quote}

\section{Discussion}

It is a common practice to assume that word representations created using the methods mentioned above can give useful information to create NLP applications. Nevertheless, upon manually analyzing all the resulting SF, we observed that some representations are ideal, ambiguous, represented in an foreign language (L2) used inside the working language (L1) and non-informative. Some examples of the last three groups include \textit{nube} (\textit{cloud}) and \textit{dominio} (\textit{domain}) from FastText; and palm from Word2Vec.

\begin{table}[h]{Different Types of Word Representations}
\small
    \centering
    \begin{tabular}{lll}
    \toprule
Non Informative & Ambiguous & L2 in L1 \\
\midrule
Dominio & Nube & Palm \\
\midrule
"Dominio" & "cloudcomputing" & "plum" \\
"dominios" & "OwnCloud" & "frog" \\
"dominio.El" & "SoftLayer" & "wood" \\
"dominio." & "ownCloud" & "leaved" \\
"eldominio" & "IaaS" & "lily" \\
"sub-dominio" & "nubecitas" & "ferns" \\
"dominio.En" & "neblina" & "oak" \\
"dominio.-" & "virtualizada" & "apple" \\
"dedominio" & "NUBE" & "maple" \\
"domino" & "SaaS" & "gum" \\
"subdominio" & "hiperconvergencia" & "found\_in" \\
"dominiode" & "niebla" & "native" \\
"dominio.La" & "clouds" & "leaf" \\
"dominio.com" & "Wordle" & "fruit" \\
"dominio-" & "vaporosa" & "jelly" \\
\bottomrule
    \end{tabular}
    \label{tab:model_eval}
\end{table}

In the case of DENISE, the use of a different method of classification ensures that the data goes trough a double-check step that turns in candidates that otherwise would be discarded. As a general recommendation, the linguistic content should be taken into account when implementing neural word embeddings. Regarding the particularities of each model, one key disadvantage of Word2Vec (specially for this task) is that it only yields representations of one meaning of the words that conform the vocabulary, and, as a consequence, there are other known meanings that are not being represented in this model. This kind of modeling could create ambiguous embeddings.

The overall performance of FastText was adequate. Even thought it might not be useful for this particular task, this kind of model might be better suited for detecting new words on a formal level, since it creates words representations for words that are not included in the vocabulary. This model could also be useful for analyzing composition and derivation processes on a lexical and morphological level.

The Sense2Vec model gave the best results for this particular task, in great part due to implementation of POS tags. These tags add information that can be used to disambiguate meaning of new words or polysemic words. However, from the 125 keywords it only had representations for 97. This might be due to the training parameters suggested by the authors or that, in comparison with FastText, we require more training data. Wikipedia is commonly used as corpus, but for a system that requires a general and broad representation of a language, more diverse data is required.

\section{Conclusions and Future Work}

In this study we have shown the application of word embeddings for the detection of semantic neologisms. For this particular task, the Sense2Vec model gave the best performance. We explored some of the advantages that FastText models have over Word2Vec; for instance, representations of uncommon words.

After further manual analysis of the most similar terms that each models generated, we observed three types of representations that are not useful for the development of the DENISE system: ambiguous embeddings, L2 in L1 embeddings and non informative embeddings. These kind of embeddings should be taken into account when designing an NLP application since the final goal is to implement rich linguistic knowledge. In the case of DENISE, we use TF-IDF and logistic regression for theme classification so the system does not rely on one single method to analyze semantic change.

\begin{figure}[h]
    \centering
    \includegraphics[scale=.5]{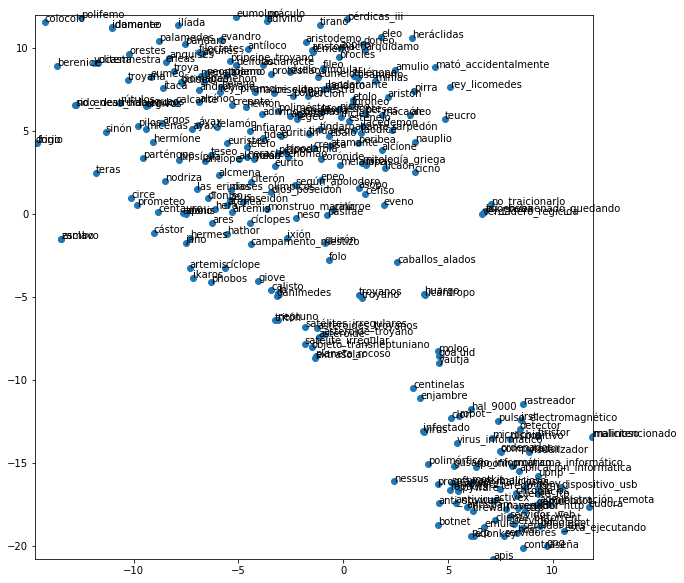}
    \caption{Most Common Words Related to ``troyano'' in Spanish}
    \label{fig:troyano}
\end{figure}

While analyzing the characteristics of the generated embeddings we could observe that ambiguous representations usually contain words related to two or more different topics. While the Sense2Vec model can differentiate between words that can be used as a verb or as a noun, this process still generates one representation per POS tag. Figure \ref{fig:troyano} shows the most common words related to \textit{troyano} (trojan in Spanish) in the Word2Vec model.DENISE classified this word as a valid candidate but, on further inspection, when selecting 300 most common words we can observe two clusters of words: one on the upper-left part that is related to its mythological sense and a small cluster on the lower-right that contains words related to the CS field.

Based on this observation, a possible future line of work might be the development of polysemic embeddings, be it as an added layer during the training process that could generate more than one representations for each word, or as a post training process using clustering or classification techniques. Such embeddings could be a good addition for other NLP related tasks such as automatic translation or Automatic Text Summarization \cite{Torres-Moreno2014,Torres-Moreno2011} or word sense disambiguation.

\section {Acknowledgements}
Authors want to thank CONACYT (\url{https://www.conacyt.gob.mx}) Convocatoria de Becas al Extranjero 2015-2019, for supporting this research.
The UILATERM group from the Pompeu Fabra University for giving access to their neologisms database and in particular Prof. Rosa Estopâ for her insights regarding the theoretical background.

\bibliographystyle{plain}
\bibliography{biblio}

\begin{thebibliography}{10}

\bibitem{Barrios2016}
Federico Barrios, Federico L{\'o}pez, Luis Argerich, and Rosa Wachenchauzer.
\newblock Variations of the {Similarity} {Function} of {TextRank} for
  {Automated} {Summarization}.
\newblock {\em CoRR}, abs/1602.0, February 2016.

\bibitem{Bernhard2015}
Delphine Bernhard, Lauren Bruneau, Ingrid Falk, and Christophe G{\'e}rard.
\newblock Cr{\'e}ation lexicale et corpus dynamique: quelles variables
  contextuelles?
\newblock In {\em 10e journ{\'e}es internationales LTT 2015 {\`a} Strasbourg:
  La cr{\'e}ation lexicale en situation: texte, genres, cultures}, Strasbourg,
  France, September 2015.

\bibitem{Bernhard2015a}
Delphine Bernhard, Lauren Bruneau, Ingrid Falk, and Christophe G{\'e}rard.
\newblock Le logoscope : une approche textuelle de la veille n{\'e}ologique.
  des raisons linguistiques {\`a} l'interface web.
\newblock In {\em CINEO 2015 - III Congreso Internacional de Neolog{\'\i}a en
  las Lenguas Rom{\'a}nicas}, Salamanca, Spain, October 2015.

\bibitem{Bojanowski2016}
Piotr Bojanowski, Edouard Grave, Armand Joulin, and Tomas Mikolov.
\newblock Enriching word vectors with subword information.
\newblock {\em arXiv preprint arXiv:1607.04606}, 2016.

\bibitem{cabre2009}
Maria~Teresa Cabr{\'{e}}.
\newblock {La classificaci{\'{o}} dels neologismes: una tasca complexa}.
\newblock In Mar{\'{\i}}a~Teresa Cabr{\'{e}} and Rosa Estop{\`{a}}, editors,
  {\em Les paraules noves: criteris per detectar i mesurar els neologismes},
  pages 11--37. Eumo Editorial, Universitat Pompeu Fabra, Vic, Barcelona, 2009.

\bibitem{Cabre2005}
Mar{\'{\i}}a~Teresa Cabr{\'{e}} and Rosa Estop{\`{a}}.
\newblock Unidades de conocimiento especializado: caracterización y
  tipología.
\newblock In Mar{\'{\i}}a~Teresa Cabr{\'{e}} and Bach Carme, editors, {\em
  Coneixement, llenguatge i discurs especialitzat}, pages 69--93. Institut
  Universitari de Lingüística Aplicada. Universitat Pompeu Fabra, Barcelona,
  2005.

\bibitem{Estopa2013}
Rosa Estop{\`{a}}.
\newblock Les aplicacions terminol{\`{o}}giques.
\newblock In Ona Dom{\`{e}}nech and Rosa Estop{\`{a}}, editors, {\em
  Llenguatges d'especialitat i terminologia}. Universitat Oberta de Catalunya,
  Barcelona, 2013.

\bibitem{Falk2014-2}
Ingrid Falk, Delphine Bernhard, and Christophe G{\'e}rard.
\newblock {De la quenelle culinaire {\`a} la quenelle politique :
  identification de changements s{\'e}mantiques {\`a} l'aide des Topic Models}.
\newblock In {\em {21{\`e}me conf{\'e}rence sur le Traitement Automatique des
  Langues Naturelles}}, Marseille, France, July 2014.

\bibitem{Falk2014-1}
Ingrid Falk, Delphine Bernhard, and Christophe G{\'e}rard.
\newblock {From Non Word to New Word: Automatically Identifying Neologisms in
  French Newspapers}.
\newblock In {\em {LREC - The 9th edition of the Language Resources and
  Evaluation Conference}}, Proceedings of the International Conference on
  Language Resources and Evaluation, Reykjavik, Iceland, May 2014.

\bibitem{Falk2014-3}
Ingrid Falk, Delphine Bernhard, Christophe G{\'e}rard, and Romain Potier-Ferry.
\newblock {{\'E}tiquetage morpho-syntaxique pour des mots nouveaux}.
\newblock In {\em {21{\`e}me conf{\'e}rence sur le Traitement Automatique des
  Langues Naturelles}}, Marseille, France, July 2014.

\bibitem{Gerard2014}
Christophe G{\'e}rard, Ingrid Falk, and Delphine Bernhard.
\newblock {Traitement automatis{\'e} de la n{\'e}ologie : pourquoi et comment
  int{\'e}grer l'analyse th{\'e}matique ?}
\newblock In {\em {Actes du 4e Congr{\`e}s Mondial de Linguistique Fran{\c
  c}aise (CMLF 2014)}}, volume~8 of {\em SHS Web of Conferences.}, pages 2627
  -- 2646, Berlin, Germany, July 2014.

\bibitem{Janssen2005}
Maarten Janssen.
\newblock {NeoTrack: semi-automatic neologism detection}.
\newblock In {\em APL XXI}, Porto, Portugal, 2005.

\bibitem{Janssen2009}
Maarten Janssen.
\newblock {Detecci{\'{o}}n de Neologismos: una perspectiva computacional}.
\newblock {\em Debate Terminol{\'{o}}gico}, 5(05):68--75, 2009.

\bibitem{Janssen2012}
Maarten Janssen.
\newblock {NeoTag: a POS Tagger for Grammatical Neologism Detection}.
\newblock {\em Proceedings of the Eight International Conference on Language
  Resources and Evaluation (LREC'12)}, (1):2118--2124, 2012.

\bibitem{Joulin2016}
Armand Joulin, Edouard Grave, Piotr Bojanowski, and Tomas Mikolov.
\newblock Bag of tricks for efficient text classification.
\newblock {\em arXiv preprint arXiv:1607.01759}, 2016.

\bibitem{Li2014}
Guangyi Li and Houfeng Wang.
\newblock Improved {Automatic} {Keyword} {Extraction} {Based} on {TextRank}
  {Using} {Domain} {Knowledge}.
\newblock In {\em Natural {Language} {Processing} and {Chinese} {Computing}},
  pages 403--413. 2014.

\bibitem{Mihalcea2004}
Rada Mihalcea and Paul Tarau.
\newblock {TextRank}: {Bringing} {Order} into {Text}.
\newblock In {\em Proceedings of the 2004 {Conference} on {Empirical} {Methods}
  in {Natural} {Language} {Processing}}, 2004.

\bibitem{Mikolov2013}
Tomas Mikolov, Kai Chen, Greg Corrado, and Jeffrey Dean.
\newblock Efficient {Estimation} of {Word} {Representations} in {Vector}
  {Space}.
\newblock {\em CoRR}, abs/1301.3, 2013.

\bibitem{Mikolov2013b}
Tomas Mikolov, Ilya Sutskever, Kai Chen, Greg Corrado, and Jeffrey Dean.
\newblock Distributed representations of words and phrases and their
  compositionality.
\newblock {\em CoRR}, abs/1310.4546, 2013.

\bibitem{Nazar2011}
Rogelio Nazar.
\newblock {Neolog{\'{\i}}a sem{\'{a}}ntica: un enfoque desde la
  ling{\"{u}}{\'{\i}}stica cuantitativa}, 2011.

\bibitem{Nazar2013}
Rogelio Nazar.
\newblock Word sense discrimination using statistic analysis of texts.
\newblock {\em Barcelona Investigaci{\'o}n Arte Creaci{\'o}n}, 1(1):5--26,
  2013.

\bibitem{Nazar2014}
Rogelio Nazar.
\newblock Una metodolog{\'\i}a para depurar los resultados de los extractores
  de t{\'e}rminos, 2014.

\bibitem{Page1999}
Lawrence Page, Sergey Brin, Rajeev Motwani, and Terry Winograd.
\newblock The pagerank citation ranking: Bringing order to the web.
\newblock Technical Report 1999-66, Stanford InfoLab, November 1999.

\bibitem{Pay2018}
Tayfun Pay, Stephen Lucci, and Jim Cox.
\newblock {An Ensemble of Automatic Keyphrase Extractors: TextRank, RAKE and
  TAKE}.
\newblock {\em Computación y Sistemas}, 23(3):703--710, 2019.

\bibitem{Renouf1998}
Antoinette Renouf.
\newblock {\em Explorations in Corpus Linguistics}, chapter Aviating among the
  Hapax Legomena: morphological grammaticalisation in current British Newspaper
  English.
\newblock Rodopi, 1998.

\bibitem{Renouf2010}
Antoinette Renouf.
\newblock {Identification automatique de la n{\'{e}}ologie lexicologique et
  s{\'{e}}mantique : questions soulev{\'{e}}es par notre m{\'{e}}thode}.
\newblock In M.~Teresa Cabr{\'{e}}, Ona Dom{\`{e}}nech, Rosa Estop{\`{a}},
  Judith Freixa, and Merc{\`{e}} Lorente, editors, {\em Actes del Congr{\'{e}}s
  Internacional de Neologia de les Lleng{\"{u}}es Rom{\`{a}}niques}, pages
  129--141, Barcelona, 2010. Intitut Universitari de Ling{\"{u}}{\'{\i}}stica
  Aplicada.

\bibitem{Renouf2012}
Antoinette Renouf.
\newblock {A finer definition of neology in English: the life-cycle of a word}.
\newblock {\em Corpus perspectives on patterns of lexis}, pages 177--208, 2012.

\bibitem{Reutenauer2011}
Coralie Reutenauer, Evelyne Jacquey, Sandrine Ollinger, Neologismes De, and
  Langues Romanes.
\newblock {Neologismes de sens: contribution {\`{a}} leur caracterisation dans
  un corpus autour du th {\`{a}}me de la crise financi{\`{e}}re.}, 2011.

\bibitem{Sablayrolles2012}
Jean-Fran{\c{c}}ois Sablayrolles.
\newblock {Extraction automatique et types de n{\'{e}}ologismes : une
  n{\'{e}}cessaire clarification}.
\newblock {\em Cahier de lexicologie}, 100(1):37--53, 2012.

\bibitem{Tebe2002}
Carles Teb{\'{e}}.
\newblock {Bases pour une s{\'{e}}lection de neologismes}.
\newblock In Elisabet~Sol{\'e} M.~Teresa~Cabr{\'e}, Judit~Freixa, editor, {\em
  L{\`{e}}xic i neologia}, pages 43--50. {Observatori de Neologia} and
  {Universitat Pompeu Fabra}, Barcelona, 2002.

\bibitem{Torres-Moreno2011}
Juan~Manuel Torres-Moreno.
\newblock {\em R{\'e}sum{\'e} automqtique de documents - une approche
  statistique}.
\newblock Herm{\`e}s-Lavoisier, 2011.

\bibitem{Torres-Moreno2014}
Juan-Manuel Torres-Moreno.
\newblock {\em {Automatic Text Summarization}}.
\newblock Cognitive science and knowledge management series. ISTE Ltd ; John
  Wiley \& Sons, Inc, London : Hoboken, NJ, 2014.

\bibitem{Trask2015}
Andrew Trask, Phil Michalak, and John Liu.
\newblock sense2vec - a fast and accurate method for word sense disambiguation
  in neural word embeddings.
\newblock {\em CoRR}, abs/1511.06388, 2015.

\end{thebibliography}

\end{document}